\documentclass[nohyperref]{article}
\usepackage[usenames,dvipsnames]{xcolor}
\usepackage[T1]{fontenc}
\usepackage{lmodern}
\usepackage[accepted]{icml2022}
\usepackage{microtype}
\usepackage{booktabs} 
\usepackage{graphics}
\usepackage{graphicx}
\usepackage{caption}
\usepackage{subcaption}
\usepackage{multirow}
\usepackage{url}
\usepackage{xparse}
\usepackage{amssymb}
\usepackage{xspace}
\usepackage{nicefrac}
\usepackage{microtype}
\usepackage{amsmath}
\usepackage{amsthm}
\usepackage{amsfonts}
\usepackage{enumerate}
\usepackage{setspace}
\usepackage{tabularx}
\usepackage{enumitem}
\usepackage{wrapfig}
\usepackage{xfrac}
\usepackage{tikz}
\usepackage[most]{tcolorbox}

\usepackage{hyperref}
\hypersetup{
 colorlinks=True,
 linkcolor=RoyalBlue,
 citecolor=RoyalBlue,
 urlcolor=RoyalBlue}

\usepackage{balance}

\newcommand{\mytilde}{\raise.17ex\hbox{$\scriptstyle\mathtt{\sim}$}}

\icmltitlerunning{The (Un)Surprising Effectiveness of Pre-Trained Vision Models for Control}

\begin{document}

\twocolumn[
\icmltitle{The (Un)Surprising Effectiveness of Pre-Trained Vision Models for Control}



\begin{icmlauthorlist}
\icmlauthor{Simone Parisi$^*$}{meta}
\icmlauthor{Aravind Rajeswaran$^*$}{meta}
\icmlauthor{Senthil Purushwalkam}{cmu}
\icmlauthor{Abhinav Gupta}{meta,cmu}
\end{icmlauthorlist}

\icmlaffiliation{meta}{Meta AI}
\icmlaffiliation{cmu}{Carnegie Mellon University}

\icmlcorrespondingauthor{Simone Parisi}{simone@robot-learning.de}
\icmlcorrespondingauthor{Aravind Rajeswaran}{aravraj@fb.com}

\icmlkeywords{representation learning, state representations, reinforcement learning, imitation learning}

\vskip 0.3in
]



\printAffiliationsAndNotice{\icmlEqualContribution
} 

\begin{abstract}
Recent years have seen the emergence of pre-trained representations as a powerful abstraction for AI applications in computer vision, natural language, and speech. However, policy learning for control is still dominated by a tabula-rasa learning paradigm, with visuo-motor policies often trained from scratch using data from deployment environments.
In this context, we revisit and study the role of pre-trained visual representations for control, and in particular representations trained on large-scale computer vision datasets. 
Through extensive empirical evaluation in diverse control domains (Habitat, DeepMind Control, Adroit, Franka Kitchen), we isolate and study the importance of different representation training methods, data augmentations, and feature hierarchies.
Overall, we find that pre-trained visual representations can be competitive or even better than ground-truth state representations to train control policies. This is in spite of using only out-of-domain data from standard vision datasets, without any in-domain data from the deployment environments.
Source code and more at \url{https://sites.google.com/view/pvr-control}.
\end{abstract}

\section{Introduction}
\label{sec:intro}

\begin{figure}[t!]
    \centering
    \includegraphics[width=\linewidth]{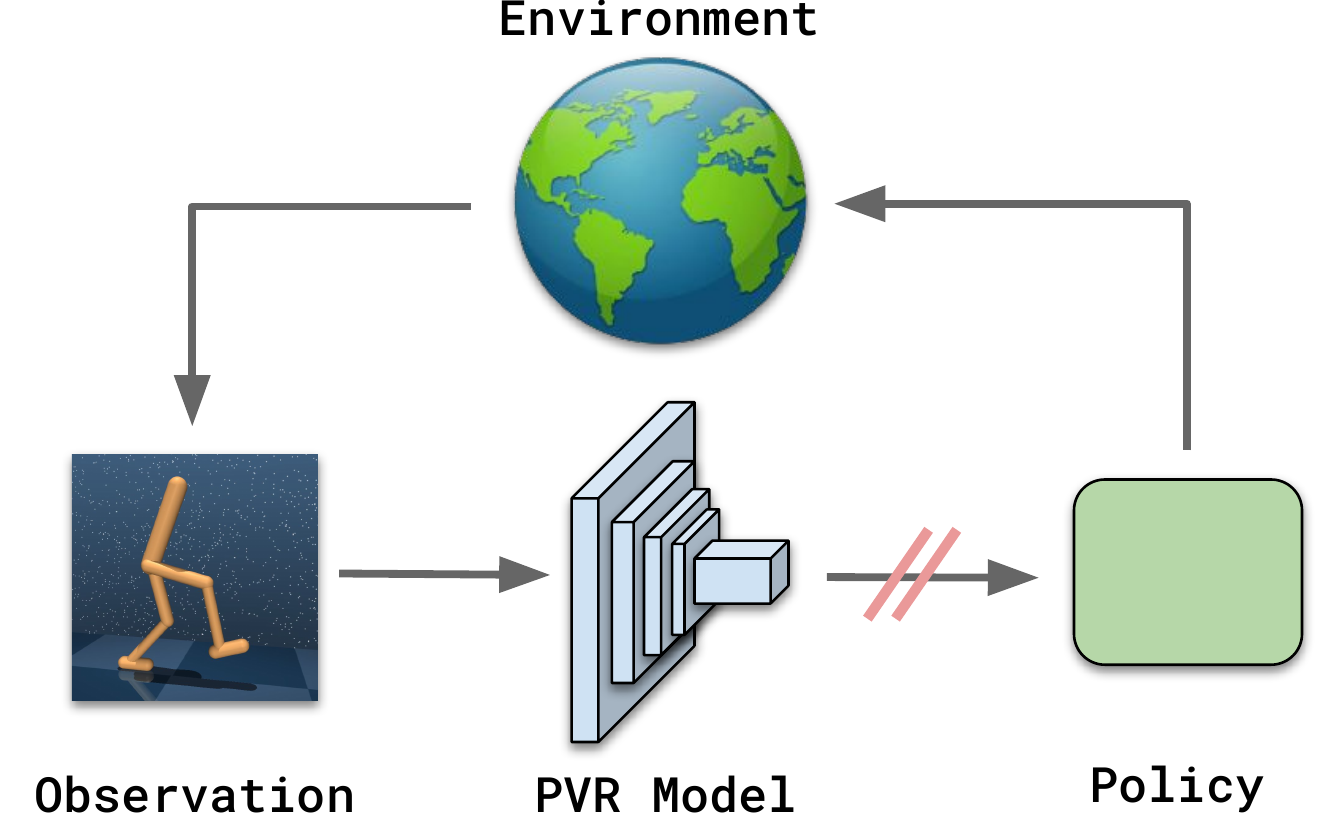} \\[8pt]
    \includegraphics[width=\columnwidth]{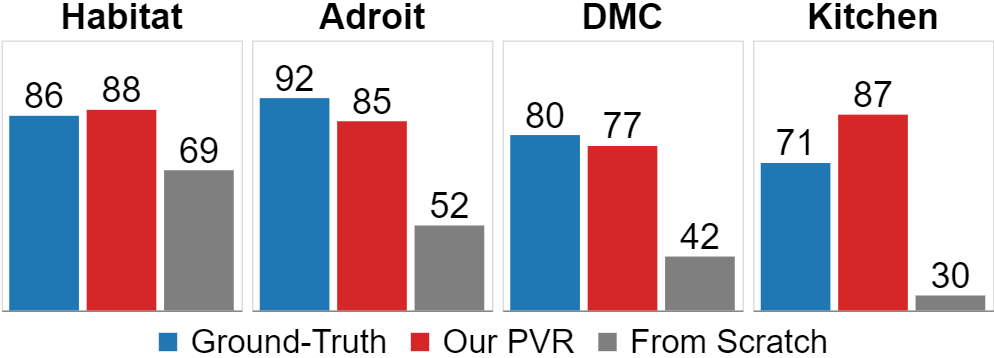}
    \caption{\label{fig:teaser}
    \textbf{(Top)} In our paradigm, a pre-trained vision model is used as a perception module for the policy. The model is frozen and  not further trained during policy updates. Its output, namely the pre-trained visual representation (PVR), serves as state representation and policy input.
    \textbf{(Bottom)} Our PVR is competitive with ground-truth features for training policies with imitation learning, in spite of being pre-trained on out-of-domain data. By contrast, the classic approach of training an end-to-end visuo-motor policy from scratch fails with the same amount of imitation data.
    }
    \vspace*{-10pt}
\end{figure}

\begin{figure*}
    \centering
    \includegraphics[width=\textwidth]{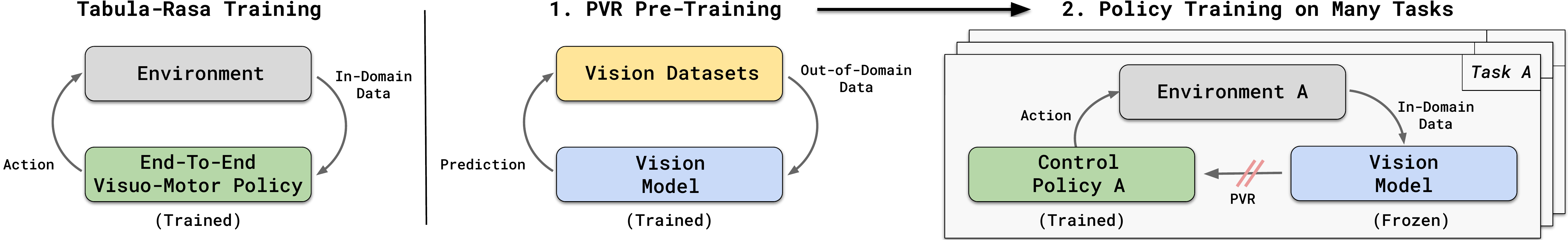}
    \caption{\textbf{Classic training paradigm (left) vs. ours (right)}. In tabula-rasa training, the perception module is part of the control policy and is trained from scratch on data from the environment. By contrast, in our paradigm the perception module is detached from the policy. First, it is trained once on out-of-domain data (e.g., ImageNet) and frozen. Then, given some tasks, control policies are trained on the deployment environments re-using the same frozen perception module.
    }
    \label{fig:our_paradigm}
\end{figure*}

Representation learning has emerged as a key component in the success of deep learning for computer vision, natural language processing~(NLP), and speech processing. Representations trained using massive amounts of labeled~\citep{krizhevsky2012imagenet,sun2017revisiting,Brown2020GPT3} or unlabeled~\citep{Devlin2019BERTPO,goyal2021self} data have been used ``off-the-shelf'' for many downstream applications, resulting in a simple, effective, and data-efficient paradigm. 
By contrast, policy learning for control is still dominated by a ``tabula-rasa'' paradigm where an agent performs millions or even billions of interactions with an environment to learn task-specific visuo-motor policies from scratch~\citep{espeholt2018impala, Wijmans2020DDPPO,yarats2021image}.

In this paper, we take a step back and ask the following fundamental question. Why have pre-trained visual representations, like those trained on ImageNet, not found widespread success in control despite their ubiquitous usage in computer vision?
Is it because control tasks are too different from vision tasks? Or because of the domain gap in the visual characteristics? Or is it that ``the devil lies in the details'', and we are failing to consider some key components?
We note that dataset domain gap is not a core issue in computer vision. For instance, ImageNet-trained models have been shown to transfer to a variety of different tasks like human pose estimation~\citep{cao2017realtime}. In this context, we aim to investigate the following fundamental question.
\begin{center}
\vspace*{-4pt}
\emph{Can we make a single vision model, pre-trained entirely on out-of-domain datasets, work for different control tasks?}
\vspace*{-4pt}
\end{center}

To answer this question, we consider a large collection of pre-trained visual representation (PVR) models commonly used in computer vision, and investigate how such models can be used as frozen perception modules for control tasks, as depicted in Figure~\ref{fig:our_paradigm}. We perform a series of experiments to understand the effectiveness of these representations in four well-known domains that require visuo-motor control policies: Habitat~\citep{habitat19iccv}, DeepMind Control~\citep{Tassa2018DeepMindCS}, Adroit dexterous manipulation~\citep{Rajeswaran-RSS-18}, and Franka kitchen~\citep{Gupta2019RelayPL}. Our investigation reveals very surprising results\footnote{We argue that our findings are surprising in the context of representation learning for control. At the same time, the success of PVRs should have been unsurprising considering their widespread success and use in computer vision.} that can be summarized as follows.

\vspace*{-5pt}
\begin{itemize}[leftmargin=*]
    \itemsep0em
    \item Our main finding is that frozen PVRs trained on completely out-of-domain datasets can be competitive with or even outperform ground-truth state features for training policies (with imitation learning). We emphasize that these vision models have never seen even a single frame from our evaluation environments during pre-training.
    \item Self-supervised learning~(SSL) provides better features for control policies compared to supervised learning.
    \item Crop augmentations appear to be more important in SSL for control compared to color augmentations. This is consistent with prior work that studies representation learning in conjunction with policy learning~\citep{Srinivas2020CURL, yarats2021image}.
    \item Early convolution layer features are better for fine-grained control tasks (MuJoCo) while later convolution layer features are better for semantic tasks (Habitat ImageNav).
    \item By combining features from multiple layers of a pre-trained vision model, we propose a single PVR that is competitive with or outperform ground-truth state features in \textbf{all} the domains we study.
\end{itemize}


\section{Related Work}
\label{sec:related_work}

\begin{figure*}[t]
\captionsetup[subfigure]{labelformat=empty}
\captionsetup[subfigure]{aboveskip=3pt,belowskip=-2pt}
\centering
\begin{subfigure}[t]{0.19\linewidth}
\includegraphics[width=\linewidth]{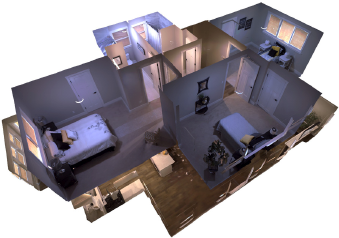}
\caption{\label{fig:habitat_apt0}Apartment 0}
\end{subfigure}
\hfill
\begin{subfigure}[t]{0.19\linewidth}
\includegraphics[width=\linewidth]{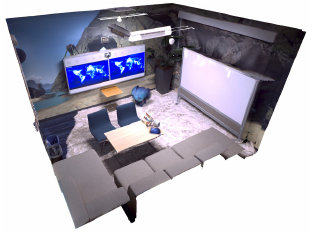}
\caption{\label{fig:habitat_apt2}Office 0}
\hfill
\end{subfigure}
\begin{subfigure}[t]{0.19\linewidth}
\includegraphics[width=\linewidth]{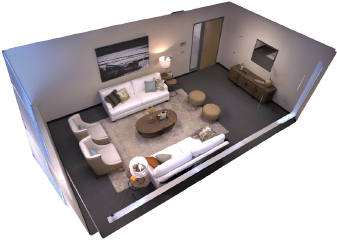}
\caption{\label{fig:habitat_room0}Room 0}
\end{subfigure}
\hfill
\begin{subfigure}[t]{0.19\linewidth}
\includegraphics[width=\linewidth]{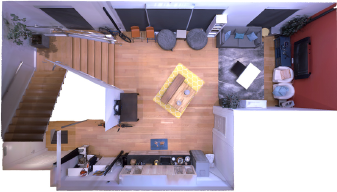}
\caption{\label{fig:habitat_frl_apt0}FRL Apartment 0}
\end{subfigure}
\hfill
\begin{subfigure}[t]{0.19\linewidth}
\includegraphics[width=\linewidth]{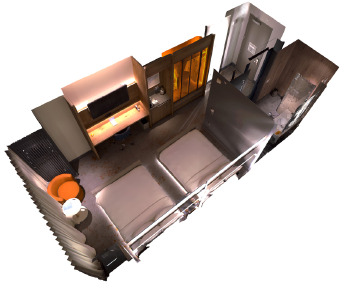}
\caption{\label{fig:habitat_hotel0}Hotel 0}
\end{subfigure}
\vspace*{-8pt}
\caption{\label{fig:replica_scenes}\textbf{Real-world scenes from the Replica dataset used in Habitat.} The agent has to reach target locations from anywhere on the scene. Its perception is based on its egocentric view of the scene and an image showing the target location. Only ground-truth state features explicitly inform the agent about its position, the target coordinates, and the scene it is in.} 
\end{figure*}

\begin{figure*}[t]
\captionsetup[subfigure]{labelformat=empty}
\captionsetup[subfigure]{aboveskip=3pt,belowskip=-2pt}
\centering
\begin{subfigure}[t]{0.130\linewidth}
\includegraphics[width=\linewidth]{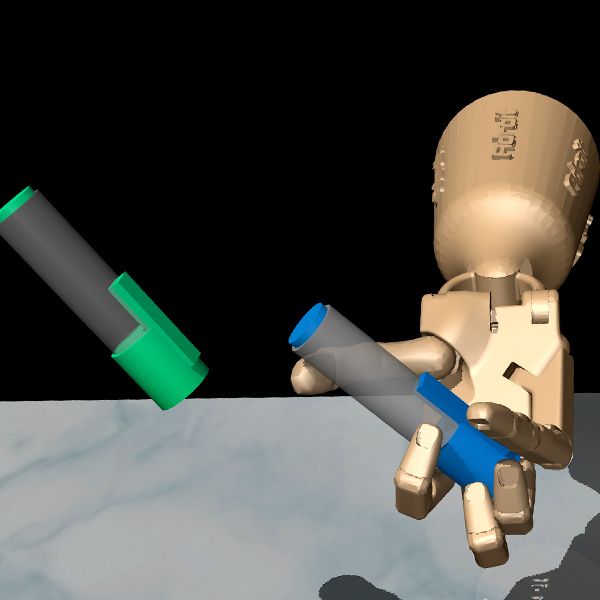}
\caption{\label{fig:adroit_pen}Adroit Pen}
\end{subfigure}
\hfill
\begin{subfigure}[t]{0.130\linewidth}
\includegraphics[width=\linewidth]{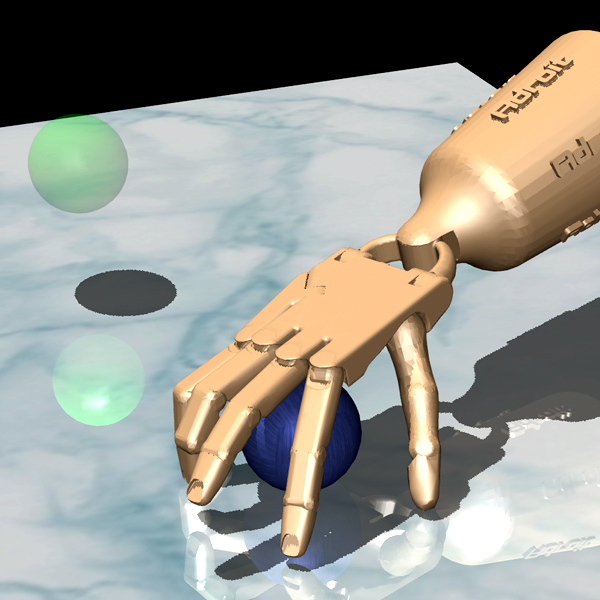}
\caption{\label{fig:adroit_relo}Adroit Relocate}
\hfill
\end{subfigure}
\begin{subfigure}[t]{0.130\linewidth}
\includegraphics[width=\linewidth]{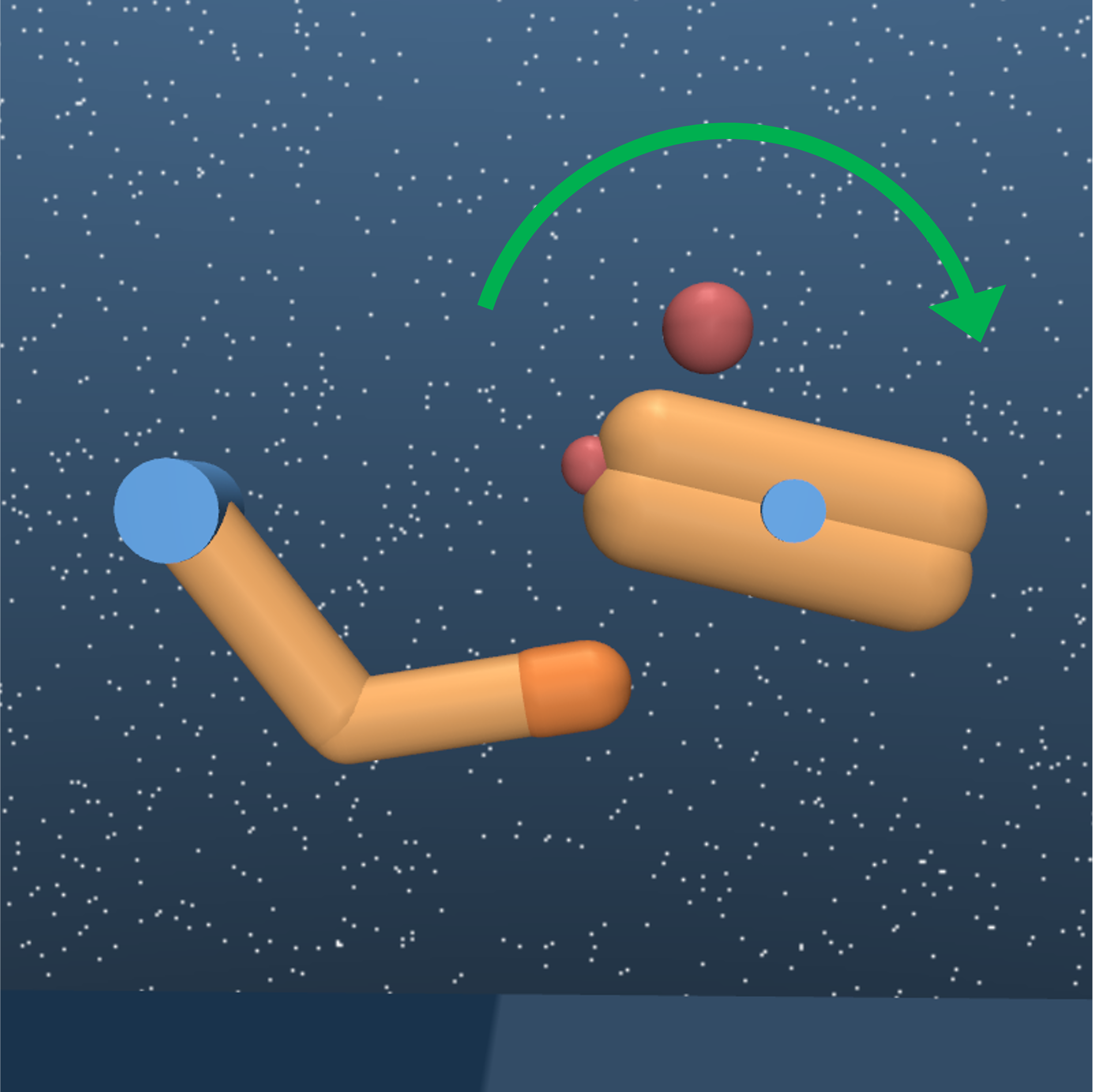}
\caption{\label{fig:cart_pole}DMC Finger Spin}
\end{subfigure}
\hfill
\begin{subfigure}[t]{0.130\linewidth}
\includegraphics[width=\linewidth]{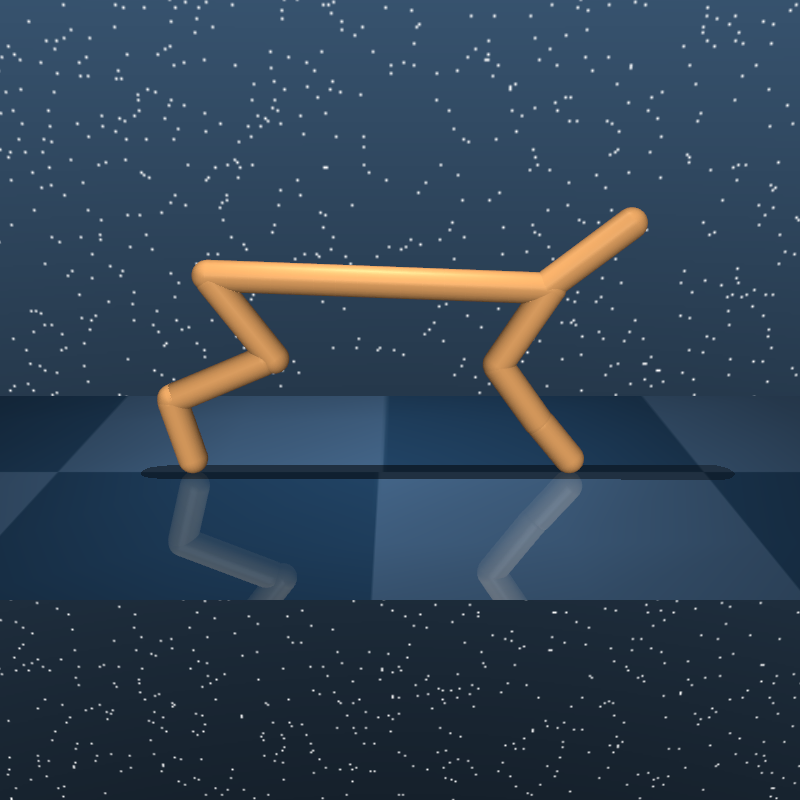}
\caption{\label{fig:cheetah}DMC Cheetah}
\end{subfigure}
\hfill
\begin{subfigure}[t]{0.130\linewidth}
\includegraphics[width=\linewidth]{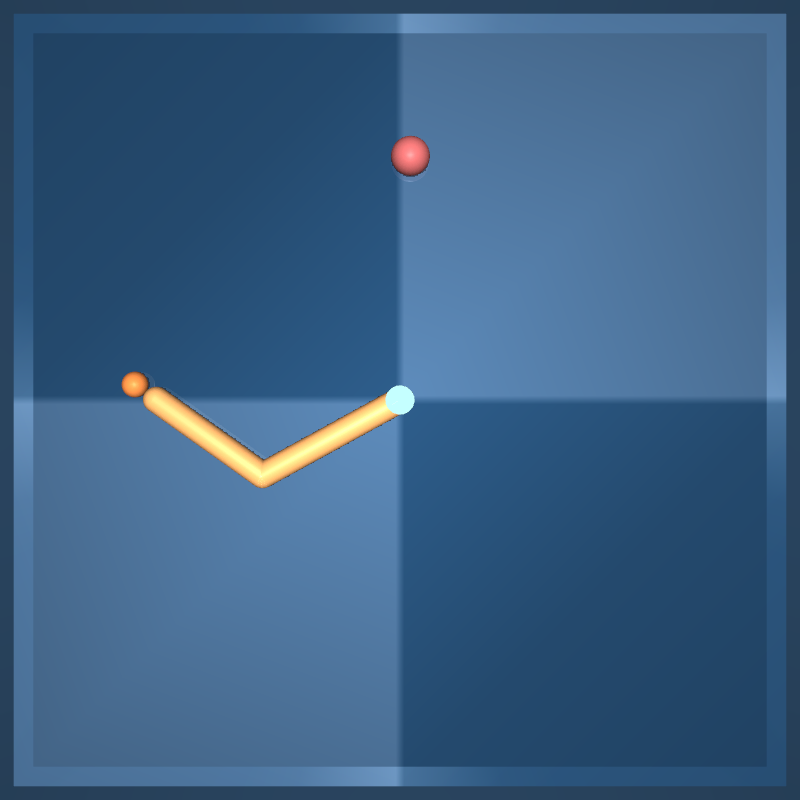}
\caption{\label{fig:reacher}DMC Reacher}
\end{subfigure}
\hfill
\begin{subfigure}[t]{0.130\linewidth}
\includegraphics[width=\linewidth]{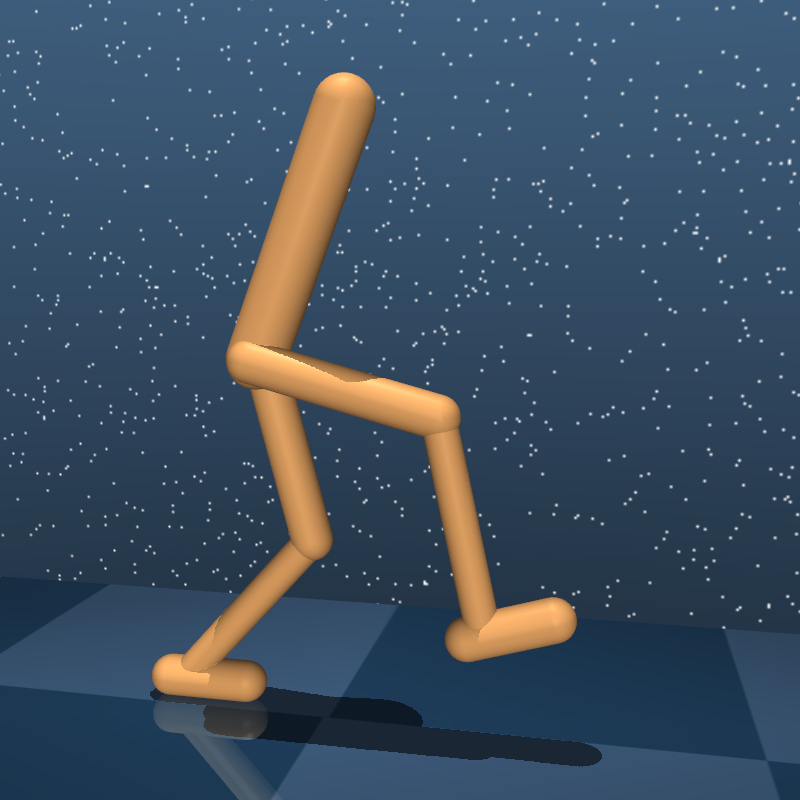}
\caption{\label{fig:walker}DMC Walker}
\end{subfigure} 
\hfill
\begin{subfigure}[t]{0.175\linewidth}
\includegraphics[width=\linewidth]{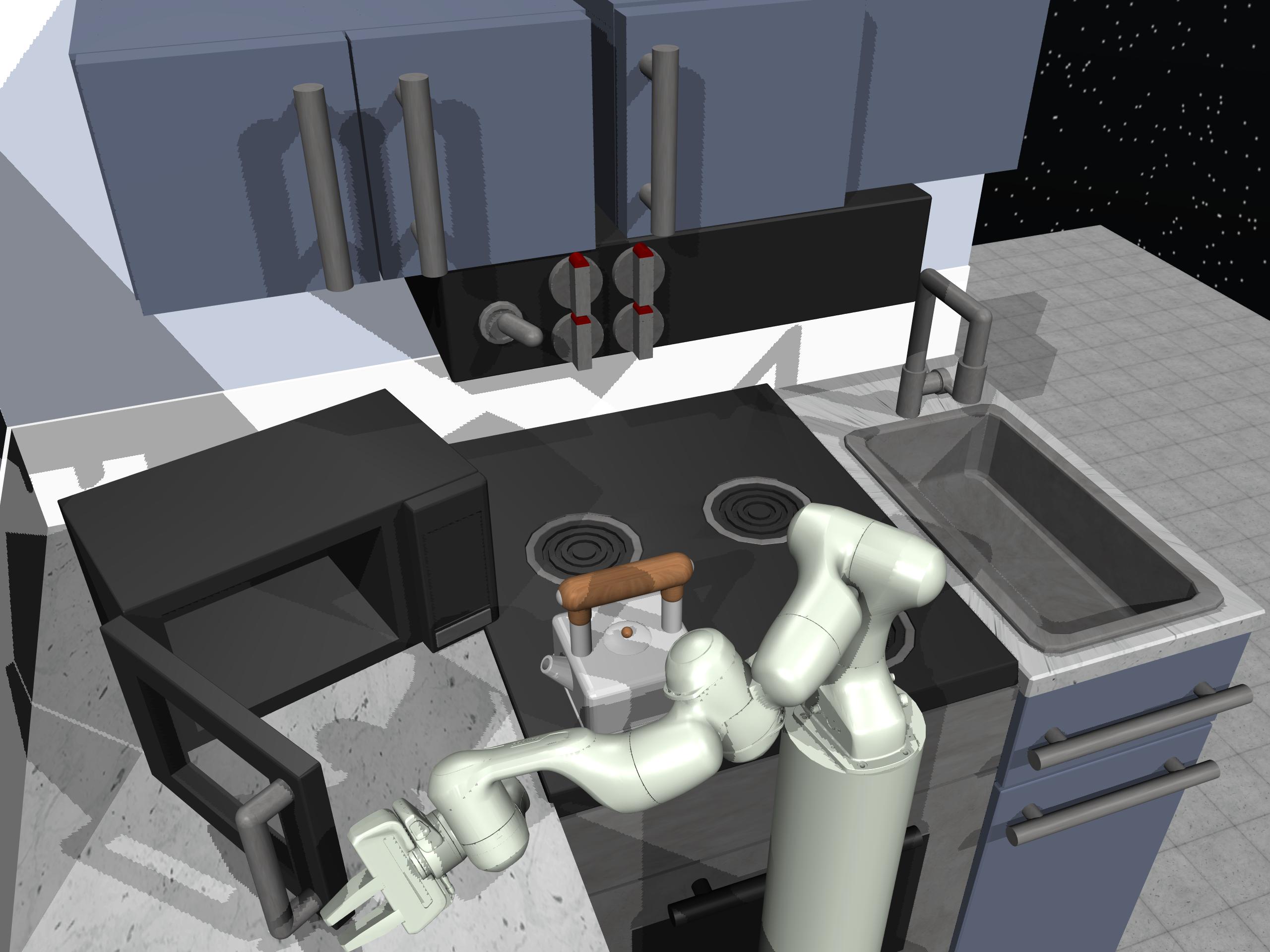}
\caption{\label{fig:kitchen}Franka Kitchen}
\end{subfigure}
\vspace*{-10pt}
\caption{\label{fig:mujoco_task_illustrations}\textbf{MuJoCo tasks span three domains.} 
In Adroit (left), the agent has to learn dexterous hand manipulation behaviors like grasping and in-hand manipulation. In the DeepMind Control suite (center), it needs to learn low-level locomotion and manipulation behaviors. In Franka Kitchen (right), it has to reconfigure objects in a kitchen using a Franka arm.
}
\end{figure*}

\textbf{Representation Learning.}
Pre-training representations and transfering them to downstream applications is an old and vibrant area of research in AI~\citep{hinton2006reducing,krizhevsky2012imagenet}.
This approach gained renewed interest in the fields of computer vision, speech, and NLP with the observation that representations learned by deep networks transfer remarkably well to downstream tasks~\citep{girshick2014rich,Devlin2019BERTPO,Baevski2020wav2vec2A}, 
resulting in improved data efficiency and/or performance~\citep{goyal2019scaling}.

Focusing on computer vision, representations can be learned either through supervised methods, such as ImageNet classification~\citep{krizhevsky2012imagenet,Russakovsky2015ImageNetLS}, or through self-supervised methods that do not require any labels~\citep{doersch2015unsupervised,Chen2020SimCLR,purushwalkam2020demystifying}.
The learned representations can be used ``off-the-shelf'', with the representation network frozen and not adapted to downstream tasks. This approach has been successfully used in object detection~\citep{girshick2014rich,girshick2015fast}, segmentation~\citep{he2017mask}, captioning~\citep{vinyals2016show}, and action recognition~\citep{hara2018can}. In this work, we investigate if frozen pre-trained visual representations can also be used for policy learning in control tasks.

\textbf{Policy Learning.} 
Reinforcement learning (RL)~\citep{sutton1998reinforcement} 
and imitation learning (IL)~\citep{abbeel2004apprenticeship}
are two popular classes of approaches for policy learning. In conjunction with neural network policies, they have demonstrated impressive results in a wide variety of control tasks spanning locomotion, whole arm manipulation, dexterous hand manipulation, and indoor navigation~\citep{Heess2017EmergenceOL, Rajeswaran-RSS-18, Peng2018DeepMimicED, Wijmans2020DDPPO, OpenAIHand, Weihs2021VisualRR}.

In this work, we focus on learning visuo-motor policies using IL. A large body of work in IL and RL for continuous control has focused primarily on learning from ground-truth state features~\citep{schulman2015trust, lillicrap2015continuous, GAIL2016Ho}. While such privileged state information may be available in simulation or motion capture systems, it is seldom available in real-world settings. This has motivated researchers to investigate continuous control from visual inputs by building upon ideas like data augmentations~\citep{Laskin2020RAD,yarats2021image}, contrastive learning~\citep{Srinivas2020CURL, Zhang2021LearningIR}, or predictive world models~\citep{Hafner2020DreamTC, Rafailov2021VMAIL}. However, these works still learn representations from scratch using frames from the deployment environments.

\textbf{Pre-trained Visual Encoders in Control.}
The use of pre-trained vision models in control tasks has received limited attention. \citet{Stooke2021DecouplingRL} pre-trained representations in DeepMind Control suite and evaluated downstream policy learning in the same domain. By contrast, we study the use of representations learned using out-of-domain datasets, which is a more scalable paradigm that is not limited by frames from the deployment environment. 
\citet{Khandelwal2021SimpleBE} studied the use of CLIP representations for visual navigation tasks and reported improved results over encoders trained from scratch. 
Similarly, \citet{yen2020learning} found that using pre-trained ResNet embeddings can improve generalization and sample efficiency for manipulation tasks, provided that the parts of the model to transfer are carefully selected.
On the other hand, \citet{shah2021rrl} reported mixed performance for pre-trained ResNet embeddings, with promising results in Adroit but negative results in DeepMind Control suite. 
Compared to these works, our study is more exhaustive: it spans four visually diverse domains, a larger collection of pre-trained representations, and different forms of visual invariances stemming from augmentations and layers. Ultimately, we find that a single pre-trained representation can be successful for all the domains we study despite their visual and task-level diversity.


\section{Experiments Setup}
\label{sec:setup}

\subsection{Environments}
\label{subsec:envs}

\textbf{Habitat}~\citep{habitat19iccv} is a home assistant robotics simulator showcasing the generality of our paradigm to a visually realistic domain. The agent is trained to navigate the five Replica scenes~\citep{replica19arxiv} shown in Figure~\ref{fig:replica_scenes}. We consider the \texttt{ImageNav} task, where the agent is given two images at each timestep corresponding to the agent's current view and the target location.

\textbf{DeepMind Control (DMC) Suite}~\citep{Tassa2018DeepMindCS} is a collection of environments simulated in MuJoCo~\citep{todorov2012mujoco}, and a widely studied benchmark in continuous control. In our evaluation, we consider five tasks from the suite: \texttt{Finger-Spin}, \texttt{Reacher-Hard}, \texttt{Cheetah-Run}, \texttt{Walker-Stand}, and \texttt{Walker-Walk}. These tasks are illustrated in Figure~\ref{fig:mujoco_task_illustrations} and require the agent to learn low-level locomotion and manipulation skills.

\textbf{Adroit}~\citep{Rajeswaran-RSS-18} is a suite of tasks where the agent must control a 28-DoF anthropomorphic hand to perform a variety of dexterous tasks. We study the two hardest tasks from this suite: \texttt{Relocate} and \texttt{Reorient Pen}, depicted in Figure~\ref{fig:mujoco_task_illustrations}. The policy is required to perform goal-conditioned behaviors where the goals (e.g., desired location/orientation for the object) has to be inferred from the scene. These environments are also simulated in MuJoCo, and are known to be particularly challenging.

\textbf{Franka Kitchen}~\citep{Gupta2019RelayPL} requires to control a simulated Franka arm to perform various tasks in a kitchen scene. In this domain, we consider five tasks: \texttt{Microwave}, \ \texttt{Left-Door}, \ \texttt{Right-Door}, \ \texttt{Sliding-Door}, and \texttt{Knob-On}. Consistent with use in other benchmarks like D4RL~\citep{fu2020d4rl}, we randomize the pose of the arm at the start of each episode, but not the scene itself.

\begin{figure}[t!]
    \centering
    \includegraphics[width=\linewidth]{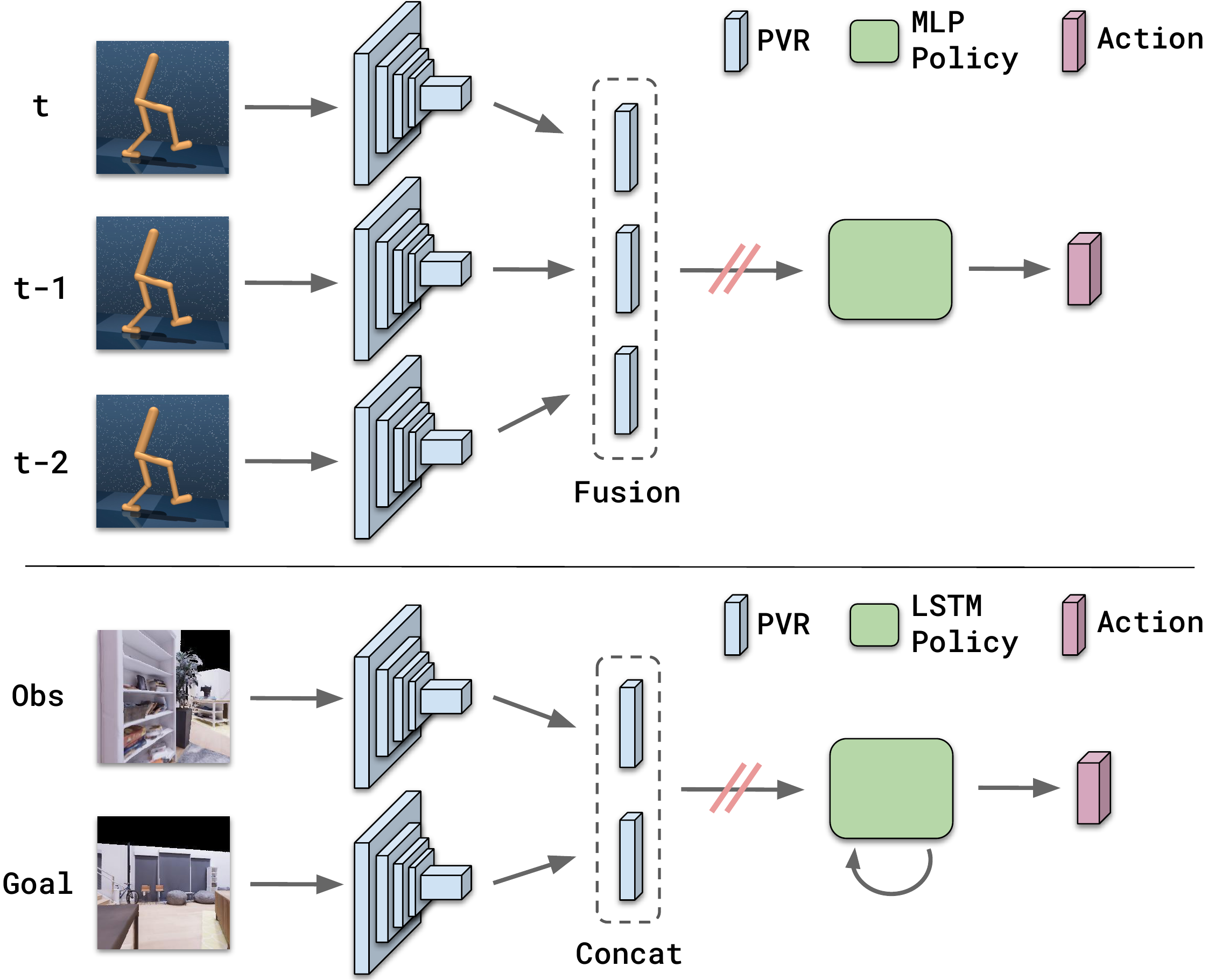}
    \caption{\label{fig:training_setup}
    \textbf{Learning architecture for MuJoCo (top) and Habitat (bottom)}. In MuJoCo, we embed the last three image observations. The resulting PVRs are then fused~\citep{Shang2021Flare} and passed to the control policy. In Habitat, we embed two images --the agent's current view of the scene and the view of the target location. The PVR embeddings are concatenated and passed to the control policy. 
    }
    \vspace*{-20pt}
\end{figure}

\begin{figure*}[t]
\centering
\includegraphics[width=\linewidth]{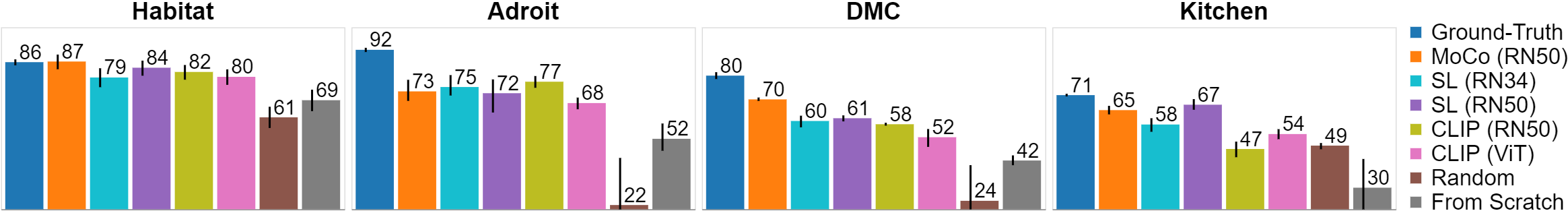}
\caption{\label{fig:pvr}\textbf{Success rate of off-the-shelf PVRs}. 
Numbers at the top of the bar report mean values over five seeds, while thin black lines denote 95\% confidence intervals.
{SL} refers to standard supervised learning as in~\cite{he2016deep}.
Any PVR is better than training the perception end-to-end from scratch together with the control policy.
In Habitat, MoCo matches the performance of ground-truth features. On the contrary, in MuJoCo, no off-the-shelf PVR can match ground-truth features.
}
\vspace*{-10pt}
\end{figure*}

\subsection{Models}
\label{subsec:models}
We investigate the efficacy of PVRs learned using a variety of models and methods including approaches that rely on supervised learning (SL) and self-supervised learning (SSL).

\textbf{Residual Network}~\citep{he2016deep} is a class of models commonly used in computer vision. Recently, ResNets have also been used in control policies, either frozen~\cite{shah2021rrl}, partially fine-tuned~\cite{Khandelwal2021SimpleBE}, or fully fine-tuned~\cite{Wijmans2020DDPPO}. In our experiments, SL (RN34) and SL (RN50) refer to ResNet-34 and ResNet-50 trained with SL on ImageNet.
\\[3pt]
\textbf{Momentum Contrast (MoCo)}~\citep{He2020MoCo} is a SSL method relying on the instance discrimination task to learn representations. These representations have shown competitive performance on many computer vision downstream tasks like image classification, object detection, and instance segmentation. MoCo uses data augmentations like cropping, horizontal flipping, and color jitter to synthesize multiple views of a single image. In our experiments, we use the pre-trained ResNet-50 model from the official repository.

\textbf{Contrastive Language-Image Pretraining (CLIP)}~\citep{radford2021learning} jointly trains a visual and textual representation using a collection of image-text pairs from the web. The learned representation has demonstrated impressive semantic discriminative power, zero-shot learning capabilities, and generalization across numerous domains of visual data. In our experiments, we use the ResNet-50 and ViT networks pre-trained with CLIP from the official repository.

\textbf{Random Features.} As baseline, we consider a randomly initialized convolutional neural network. 
Similarly to previous models, this network is frozen and not updated during learning. For the architecture details, we refer to Appendix~\ref{supp:details}.

\textbf{From Scratch.} We also compare with the classic end-to-end approach, where the aforementioned random convolutional network is trained as part of the policy. We argue that this is an inefficient approach to train visuo-motor policies, as learning good visual encoders is known to be data-hungry.

\textbf{Ground-Truth Features.} 
These are compact features provided by the simulator, and describe the full state of the agent and environment. Because in real-world settings the state can be hard to estimate, we can view these features as an ``oracle'' baseline that we strive to compete with.

\subsection{Policy Learning and Evaluation with PVRs}
\label{subsec:eval}
After pre-training, the aforementioned models are frozen and used as a perception module for the control policy. The policy is trained by IL (specifically, behavioral cloning) over optimal trajectories, and its success is estimated using evaluation rollouts in the environments.
\vspace*{-4pt}
\begin{itemize}[leftmargin=*]
\itemsep0em
\item In Habitat, training trajectories are generated using its native solver that returns the shortest path between two locations.
We collect 10,000 trajectories per scene, for a total of \mytilde2.1 million data points.
A policy is successful if the agent reaches the destination within the steps limit.
\item In MuJoCo, training trajectories are collected using a state-based optimal policy trained with RL. We collect between 25-100 trajectories per task, depending on our estimate of the task difficulty. For Adroit and Kitchen, we report the policy success percentage provided by the environments. For DMC, we report the policy return rescaled to be in the range of $[0, 100]$.
\end{itemize}

The learning setup is summarized in Figure~\ref{fig:training_setup}. In line with standard design choices, we use an LSTM policy to incorporate trajectory history in Habitat~\citep{Wijmans2020DDPPO,parisi2021interesting}, and an MLP with fixed history window in MuJoCo~\citep{yarats2021image, Laskin2020RAD}. 


\begin{figure*}[t]
\setlength{\abovecaptionskip}{3pt}
\centering
\includegraphics[width=0.99\linewidth]{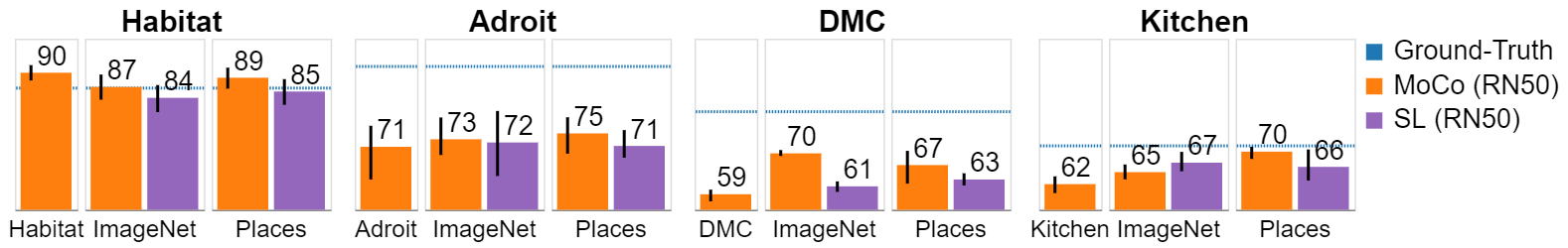}
\caption{\label{fig:datasets}\textbf{In-domain vs. out-of-domain training datasets.} 
Training PVRs on in-domain data does not help achieving better performance. In MuJoCo it even worsen it. If not the domain gap, what is the primary reason of PVRs failures?
}
\end{figure*}


\section{Experiments Results and Discussion}
\label{sec:results}
In the previous sections, we explained the experimental setup for training control policies using behavior cloning, and the testing environments from Habitat and MuJoCo. In this section, we experimentally study the performance of PVRs outlined in Section~\ref{sec:setup}. In particular, we study how well these representations perform out of the box, and how we could potentially improve or customize them, with the ultimate goal of better understanding the relationship between visual perception and control policies. 
For hyperparameter details see Appendix~\ref{supp:details}. For source code visit \url{https://sites.google.com/view/pvr-control}.

\subsection{How do Off-the-Shelf Models Perform for Control?} 
\label{subsec:exp_q1}
We first study how the pre-trained vision models presented in Section~\ref{subsec:models} perform off-the-shelf for our control task suite. That is, we download these models --pre-trained on ImageNet~\cite{Deng2009ImageNetAL}-- and pass their output as representations to the control policy.
The results are summarized in Figure~\ref{fig:pvr}.
Firstly, we find that \textbf{any} PVR is clearly better than both frozen random features and learning the perception module from scratch, in the small-dataset regime we study. This is perhaps not too surprising, considering that representation learning is known to be data intensive.

However, Figure~\ref{fig:pvr} also provides mixed results as no PVR is clearly superior to any other across all four domains. Nonetheless, on average, SSL models (MoCo) are better than SL models (RN50, CLIP). In particular, MoCo is competitive with ground-truth features in Habitat, but no off-the-shelf PVR can match the ground-truth features in MuJoCo. Why is this so, and can we customize the PVRs to perform better for all control tasks? We investigate different hypotheses and customizations in the following sub-sections. 


\subsection{Datasets and Domain Gap}

The PVRs evaluated above were representations from vision models trained on ImageNet~\citep{Deng2009ImageNetAL}. Clearly, ImageNet's visual characteristics are very different from Habitat and MuJoCo's. Could this domain gap be the reason why PVRs are not competitive with ground-truth features in all domains?
To investigate this, we introduce new datasets for pre-training the vision models. The first is Places~\citep{zhou2017places}, another out-of-domain dataset like ImageNet commonly used in computer vision. While ImageNet is more object-centric, Places is more scene-centric as it was developed for scene recognition.
The other datasets are in-domain images from Habitat and MuJoCo, i.e., they each contain only images from the deployment environment.

For the Places dataset, we pre-train both supervised and self-supervised vision models. For the Habitat and MuJoCo datasets, we only pre-train self-supervised models since no direct supervision is available. Moreover, pre-training models using environment data (Habitat, MuJoCo) requires design decisions like data collection policy and dataset size.
For the sake of simplicity, we collect trajectories using the same expert policies used for IL.
Larger or more diverse datasets from these environments may further improve the quality of the pre-trained representations, but run contrary to the motivation of simple and data-efficient learning.

Figure~\ref{fig:datasets} summarizes the results for the aforementioned representations. While in-domain pre-training helps compared to training from scratch, it is surprisingly not much better than pre-training on ImageNet or Places. For Habitat, pre-training on Habitat leads to similar performance as pre-training on ImageNet and Places. However, in the case of MuJoCo, PVRs trained on the MuJoCo expert trajectories are not competitive with representations trained on ImageNet or Places. As mentioned earlier, training on larger and more diverse datasets \textit{may} potentially bridge the gap, but is not a pragmatic solution, since we ultimately desire data efficiency in the deployment environment.

This suggests that the key to representations that work on diverse control domains does not lie only in the training dataset. Our next hypothesis is that it perhaps lies in the invariances captured by the model. 



\begin{figure}[t]
\centering
\includegraphics[width=\linewidth]{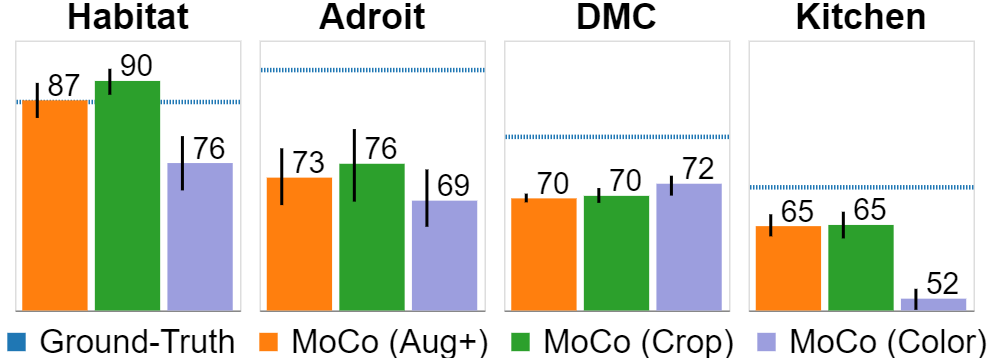}
\caption{\label{fig:invariances}\textbf{Invariances comparison in MoCo.} {Aug+} denotes the use of all augmentations as in~\cite{He2020MoCo}. 
Color-only augmentation performs worse in all environments except for DMC, while crop-only augmentation performs the best on average. This suggests that color invariance, commonly used in semantic recognition, is not always suited for control.
}
\end{figure}

\begin{figure*}[t]
\centering
\includegraphics[width=\linewidth]{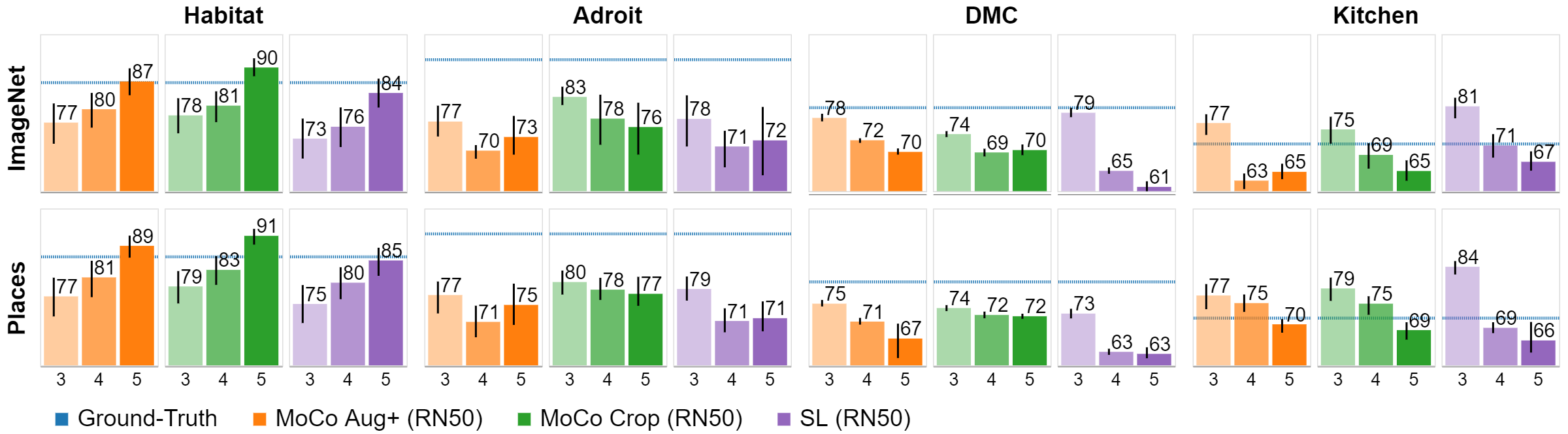}
\caption{\label{fig:layers}\textbf{Success rate when using PVRs from layers 3, 4, and 5}. There is a clear trend in Habitat showing that PVRs from later layers (opaque colors) perform better. By contrast, early layer features (transparent colors) perform better in MuJoCo. The same trends hold across both ImageNet and Places.}
\end{figure*}

\subsection{Recognition vs. Control: Two Tales of Invariances}

Most off-the-shelf vision models have been designed for semantic recognition. Next, we investigate if representations for control tasks should have different characteristics than representations for semantic recognition. Intuitively, this does seem obvious. For example, semantic recognition requires invariances to poses/viewpoints, but poses provide critical information to action policies. To investigate this aspect, we conduct the following experiment on MoCo. By default, MoCo learns invariances through various data augmentation schemes: crop augmentation provides translation and occlusion invariance, while color jitter augmentation provides illumination and color invariance. In this experiment, we isolate such effects by training MoCo with only one augmentation at a time. In semantic recognition, both color and crop augmentations appear to be critical~\citep{Chen2020SimCLR}. Does this hold true in control as well?

Results in Figure~\ref{fig:invariances} indicate that different augmentations have dramatically different effects in control. In particular, in all domains other than DMC, color-only augmentations significantly under-perform. Furthermore, crop-only augmentations lead to representations that are as good or even better than all other representations. The importance of crop-only augmentations is consistent with prior works as well~\citep{Srinivas2020CURL,yarats2021image}. We hypothesize that crop augmentations highlight relative displacement between the agent and different objects, as opposed to their absolute spatial locations in the image observation, thus providing a useful inductive bias. Overall, our experiment suggests that control may require a different set of invariances compared to semantic understanding.

\subsection{Feature Hierarchies for Control}
\label{subsec:layers}
The previous experiment indicates that invariances for semantic recognition may not be ideal for control. So far, we have leveraged the features obtained at the last layer (after final spatial average pooling) of pre-trained models. This layer is known to encode high-level semantics~\citep{selvaraju2017grad,feng2019self}.
However, control tasks could benefit from access to a low-level representation that encodes spatial information.
Furthermore, studies in vision have shown that last layer features are the most invariant and early layer features are less invariant to low-level perturbations~\citep{Zeiler2014VisualizingAU}, which have resulted in the use of feature pyramids and hierarchies in several vision tasks~\citep{lin2017feature}. Inspired by these observations, we next investigate the use of early layer features for control. We note that intermediate layers (third, fourth) have more activations than the last layer (fifth). To ease computations and perform fair comparisons, we compress these representations to the size of the representation at the last layer (more details in Appendix~\ref{supp:compression}). To the best of our knowledge, the use of early layer features is still unexplored in policy learning for control. 


\begin{figure*}
\centering
\includegraphics[width=\linewidth]{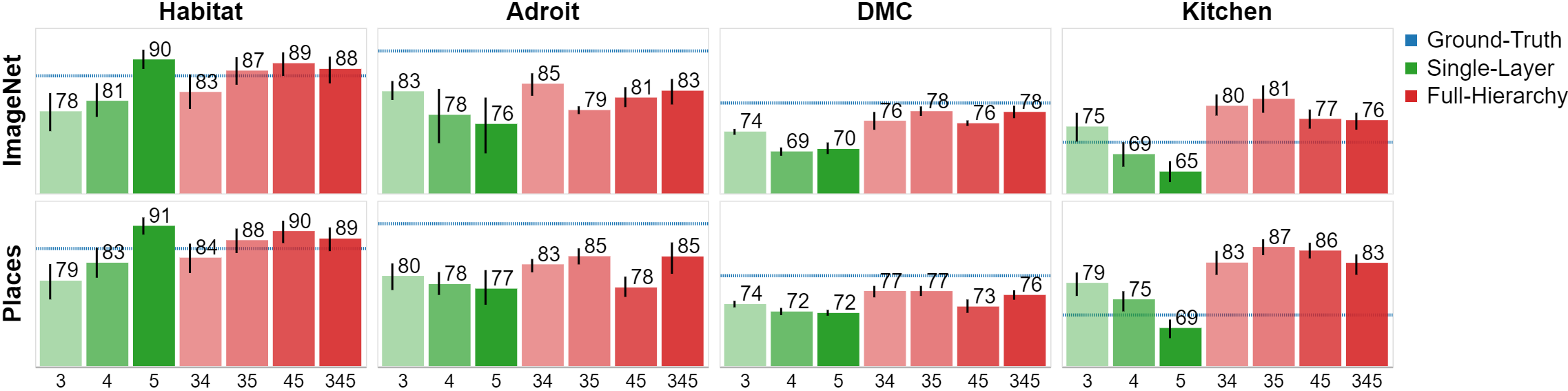}
\caption{\label{fig:uber}\textbf{Single-layer vs. full-hierarchy features of MoCo with crop-only augmentation.} The latter are competitive with ground-truth features in \textbf{all} the domains, and in the case of Kitchen even outperform them.}
\end{figure*}

Figure~\ref{fig:layers} shows that early convolution layer features are more effective for fine-grained control tasks (MuJoCo). In fact, they are so effective that they even match or outperform ground-truth features. While the ground-truth state features we use contain complete information --i.e., can function as Markov states-- they may not be the ideal representation from a learning viewpoint\footnote{We emphasize that the ground-truth features used in our experiments are the default choices provided by the environments and have been used in many prior works.}.
Indeed, not only are state features known to impact policy learning performance~\citep{brockman2016openai, Ahn2019ROBELRB}, but different representations of the same information --e.g., Euler angles and quaternions-- may perform differently~\cite{Gaudet2018DeepQN}. At the same time, visual representations may capture higher-level information that makes it easier for the agent to behave optimally.

%

Furthermore, earlier layer features work better for MuJoCo but not for Habitat. This is perhaps not surprising since navigation in Habitat requires semantic understanding of the environment. For instance, the agent needs to detect if there is a wall or an obstacle in front of itself to avoid it. This kind of information may be present in the last layer of vision model trained for semantic recognition.

\subsection{Full-Hierarchy Models}
\label{subsec:uber}

The experiment in Section~\ref{subsec:layers} motivates two new questions. First, can we design PVRs combining features from multiple layers of vision models? Ideally, the policy should learn to use the best features required to solve the task.
Second, since PVRs work even when pre-trained on out-of-domain data, could such new full-hierarchy features be ``near-universal'', i.e., work for any control task --at least those studied here?

Figure~\ref{fig:uber} shows the success of PVRs using all combinations of the last three layers of MoCo with crop-only  augmentation, the best model so far.
In MuJoCo, any PVR using the third layer features --the best single-layer features-- performs competitively with 
ground-truth features.
Similarly, in Habitat any PVR using the fifth layer performs extremely well. 
This suggests that the policy can indeed exploit the best features from the full-hierarchy to solve the task.

Overall, the PVR using all the three layers (3, 4, 5) performs best on average, and the same PVR is able to solve all the four domains, sometimes even better than ground-truth features. This is an important result, considering  
that our four control domains are very diverse and span low-level locomotion, dexterous manipulation, and indoor navigation in very diverse environments. Furthermore, this PVR is trained entirely using out-of-domain data and has never seen a single frame from any of these environments. This presents a very promising case for using PVRs for control.


\section{Discussion and Conclusion}
\label{sec:concl}

\textbf{PVR: Freezing vs. Fine-Tuning.} The prime motivation of our work is to study the use of representations from pre-trained vision models for control, and see if it is possible to develop a PVR that works in all of our testing domains. 
Consistent with this, our experiments freeze the vision models and prevent any ``on-the-fly'' representation fine-tuning. This is similar in spirit to the linear classification (probe) protocol used to evaluate representations in computer vision. We leave evaluation of representations in the full fine-tuning regime to future work.

\textbf{Imitation Learning vs. Reinforcement Learning.} In this work, we focused on learning policies using IL (specifically, behavior cloning) as opposed to RL. Despite significant advances in learning visuo-motor policies with RL~\citep{yarats2021reinforcement, Wijmans2020DDPPO, Hafner2020DreamTC}, the best algorithms are still data-intensive and require millions or billions of samples. The use of pre-trained representations are particularly important in the sparse-data regime, and thus we choose to train policies with IL. Furthermore, our work required the evaluation of a large collection of pre-trained models across many diverse environments, which was prohibitively expensive with current RL algorithms. We hope that the insights resulting from our experiments can be used to further improve RL for control in future work.

\textbf{Summary of Our Contibutions.}
The use of off-the-shelf vision models as perception modules for control policies is a relatively new area of research, trying to bridge the gap between advances in computer vision and control.
This is a departure from the current dominant paradigm in control, where visual encoders are initialized randomly and trained from scratch using environment interactions.

In this paper, we took a step back and asked fundamental questions about representations and control, in the hope of making a single off-the-shelf vision model --trained on out-of-domain datasets-- work for different control tasks.
Through extensive experiments, we find that off-the-shelf PVRs trained on completely out-of-domain data can be competitive with ground-truth features for training policies. 
Overall, we identified three major components that are crucial for successful PVRs. 
First, SSL models provide better features for control than supervised models. Second, translation and occlusion invariance, provided by crop augmentation, is more relevant for control than other invariances like illumination and color.
Third, early convolution layer features are better for fine-grained control tasks (MuJoCo) while later convolution layer features are better for semantic tasks (Habitat).

\textbf{Towards Universal Representations for Control.}
Based on these findings, we proposed a novel PVR combining features from multiple layers of a crop-augmented MoCo model trained on out-of-domain data. Our PVR 
was competitive with or outperformed ground-truth features on all four evaluation domains.

Motivated by these results, we believe that research should focus more on learning control policies directly from visual input using pre-trained perception modules, rather than using hand-designed ground-truth features. While such features may be available in simulation or specialized motion capture systems, they are hard to estimate in unstructured real-world environments. Yet, training an end-to-end visuo-motor policy has difficulties as well. The visual encoders increase the complexity of the policies, and might require a significantly larger amount of training data. 
In this context, the use of pre-trained vision modules can offer substantial benefits by dramatically reducing the data requirement and improving the policy performance. Furthermore, using a frozen PVR simplifies the control policy architecture and training pipeline.

We hope that the promising results presented in this paper will inspire our research community to focus more on developing a universal representation for control --one single PVR pre-trained on out-of-domain data that can be used as perception module for {any} control task.

\clearpage

\balance 

\bibliography{rl_bib}
\bibliographystyle{icml2022}

\clearpage

\appendix

\section{Training Details}
\label{supp:details}

\subsection{Habitat Details}
\label{supp:hyper_hab}

\textbf{Visual Input.} PVR models are fed with two 64$\times$64 RGB images, one for the view of the scene from the agent's perspective, and one for the target location. Each image is encoded independently by the model, and the two encodings are concatenated before being passed to the policy. 
\\[1pt]
\textbf{Ground-Truth Features.} Used as baseline against PVRs, it is a 12-dimensional vector composed of: agent's position and quaternion, target's position, scene's ID and version.
\\[1pt]
\textbf{Random Features.} Following~\citet{parisi2021interesting}, we use five convolutional layers, each with 32 filters, 3$\times$3 kernel, stride 2, padding 1, and ELU activation.
\\[1pt]
\textbf{Policy Architecture.} 
The PVR passes through a batch normalization layer and then through a 2-layer MLP (ReLU activation), followed by a 2-layer LSTM
and then a 1-layer MLP (softmax activation). All hidden layers have 1,024 units. Ground-truth features do not use batch-normalization, as it significantly harmed the performance. 
\\[1pt]
\textbf{Policy Optimization.} Following \citet{parisi2021interesting}, we update the policy with 16 mini-batches of 100 consecutive steps with the RMSProp optimizer~\citep{tieleman2017divide} 
(learning rate 0.0001).
Gradients are clipped to have max norm 40.
Learning lasts for 125,000 policy updates.
\\[1pt]
\textbf{Success Rate.} The policy success rate is estimated over 50 online trajectories, and further averaged over the last six policy updates, for a total of 300 trajectories per seed.
\\[1pt]
\textbf{Imitation Learning Data.} We collect 50,000 optimal trajectories (10,000 per scene) using Habitat's native solver, for a total of \mytilde2,100,000 samples.

\subsection{MuJoCo Details}
\label{supp:hyper_muj}

\textbf{Visual Input.}
Consistent with prior works, the visual input takes the last three 256$\times$256 RGB image observations of the environment. Each image is encoded independently by the PVR model. These three PVRs are fused together by using latent differences following the work of \citet{Shang2021Flare}. We \emph{do not} use any other proprioceptive observations like joint encoders for hands, and our policies are based solely on embeddings of the visual inputs.
\\[1pt]
\textbf{Ground-Truth Features.} It is a low-dimensional vector provided by the simulator, encoding information about the agent (e.g., joints position) and the environment (e.g., goal position). Its size depends on the agent and the task to be solved. For more information we refer to \citet{Tassa2018DeepMindCS,Rajeswaran-RSS-18,Gupta2019RelayPL}.
\\[1pt]
\textbf{Random Features.} 
Following \citet{Yarats2021DrQv2}, we use a 4-layer convolutional network with 32 filters in each layer, 3$\times$3 kernel, stride 1, padding 0, and ReLU activation. The network also has batch normalization and max pooling (stride 2) between each layer, and dropout with 20\% probability between layers two and three.
\\[1pt]
\textbf{Policy Architecture.}
The fused PVR passes through a batch normalization layer and then through a 3-layer MLP with $256$ hidden units each and ReLU activation.
\\[1pt]
\textbf{Policy Optimization.}
We update the policy with mini-batches of $256$ samples for $100$ epochs with the Adam optimizer~\citep{kingma2014adam} (learning rate $0.001$). The total number of policy updates varies based on the dataset size.
\\[1pt]
\textbf{Success Rate.} We evaluate the policy every two epochs over 100 online trajectories, and report the average performance over the three best epochs over the course of learning. This way we ensure that each representation is given sufficient time to learn, and that the best performance is reported. 
\\[1pt]
\textbf{Imitation Learning Data.} We collect trajectories using an expert policy trained with RL~\cite{rajeswaran2017towards,Rajeswaran-RSS-18}. The amount of data depends on 
the task difficulty.
\begin{itemize}[leftmargin=*,noitemsep,topsep=0pt]
    \item {Adroit:} $100$ trajectories per task with $100$- and $200$-step horizon for \texttt{Reorient Pen} and \texttt{Relocate}, respectively. The total number of samples is thus 10,000 and 20,000, respectively.
    \item {DeepMind Control:} $100$ trajectories per task. We use an action repeat of $2$, resulting in a $500$-step horizon per trajectory. The total number of samples is 50,000 per task.
    \item {Franka Kitchen}: $25$ trajectories per task with $50$-step horizon for all tasks. The total number of samples is 6,250 (1,250 per task).
\end{itemize}


\subsection{PVRs Details}
\label{supp:pvr_datasets}

\textbf{Datasets}
\vspace*{-4pt}
\begin{itemize}[leftmargin=*,noitemsep,topsep=0pt]
\item ImageNet: 1.2 million images.
\item Places: 1.8 million images.
\item Habitat: \mytilde2.4 million images. We collect 20,000 optimal trajectories from all the 18 Replica scenes, keeping only one frame every three for the sake of diversity. 
\item MuJoCo: we collect 30,000 images from Adroit, 250,000 from DeepMind Control, and 25,000 from the Kitchen. For Adroit and DeepMind Control, the images are taken from the same aforementioned expert trajectories used for imitation learning. For the Kitchen, we collected more trajectories with the expert policy, since the imitation learning dataset size (6,250) was too small. We stress that these additional trajectories were used only for training the PVRs, not the policy.
\end{itemize}

\vspace*{-1pt}

\textbf{Vision Models}
\vspace*{-4pt}
\begin{itemize}[leftmargin=*,noitemsep,topsep=0pt]
\item ResNet: \url{github.com/pytorch/vision}.
\item MoCo: \url{github.com/facebookresearch/moco} (v2 version).
\item CLIP: \url{github.com/openai/CLIP} (ViT-B/32 and RN50 versions).
\end{itemize}

\subsection{Intermediate Layers Compression}
\label{supp:compression}
In Section~\ref{subsec:layers} we discussed the use of features from intermediate layers of vision models. However, the number of activations in these layers (third, fourth) is significantly higher compared to the representation at the last layer (fifth). To avoid prohibitively expensive compute requirements and perform fair comparisons across layers, we compress these representations to a common size, i.e., the size of the representation at the fifth layer. This is accomplished by adding two residual blocks to the model at the chosen intermediate layer. Similar to an autoencoder model, the first residual block compresses the number of channels, while the second residual block expands the number of channels back to the original. With these additional layers randomly initialized, the model is fine-tuned on the original pre-training task. The output of the first residual block provides the compressed features which are then used in our experiments.

\subsection{Compute Details}
\label{supp:compute}
Vision models pre-training and layer compression was distributed over two nodes of a SLURM-based cluster. Each node used four NVIDIA GeForce GTX 1080 Ti GPUs. 
Pre-training one PVR model took between 1-3 days depending on the training method, size of the model, and dataset used.
\\[1pt]
Policy imitation learning was performed on a SLURM-based cluster, using a NVIDIA Quadro GP100 GPU. Training one policy took between 8-24 hours (including policy evaluation) depending on the PVR and the environment.

%

\end{document}